\documentclass[10pt,twocolumn,letterpaper]{article}

\usepackage{cvpr}
\usepackage{times}
\usepackage{epsfig}
\usepackage{graphicx}
\usepackage{amsmath}
\usepackage{amssymb}

\usepackage[breaklinks=true,bookmarks=false]{hyperref}

\cvprfinalcopy 

\def\cvprPaperID{5028} 

\begin{document}
	\title{Self-supervised Equivariant Attention Mechanism \\for Weakly Supervised Semantic Segmentation}
	
	\author{Yude Wang\textsuperscript{\rm 1,2}, Jie Zhang\textsuperscript{\rm 1,2}, Meina Kan\textsuperscript{\rm 1,2}, Shiguang Shan\textsuperscript{\rm 1,2,3}, Xilin Chen\textsuperscript{\rm 1,2}\\
		\textsuperscript{\rm 1}Key Lab of Intelligent Information Processing of Chinese Academy of Sciences (CAS),\\Institute of Computing Technology, CAS, Beijing, 100190, China\\
		\textsuperscript{\rm 2}University of Chinese Academy of Sciences, Beijing, 100049, China\\
		\textsuperscript{\rm 3}CAS Center for Excellence in Brain Science and Intelligence Technology, Shanghai, 200031, China\\
		{\tt\small yude.wang@vipl.ict.ac.cn, \{zhangjie, kanmeina, sgshan, xlchen\}@ict.ac.cn}}
	\maketitle
	
	\begin{abstract}
		Image-level weakly supervised semantic segmentation is a challenging problem that has been deeply studied in recent years. Most of advanced solutions exploit class activation map (CAM). However, CAMs can hardly serve as the object mask due to the gap between full and weak supervisions. In this paper, we propose a self-supervised equivariant attention mechanism (SEAM) to discover additional supervision and narrow the gap. Our method is based on the observation that equivariance is an implicit constraint in fully supervised semantic segmentation, whose pixel-level labels take the same spatial transformation as the input images during data augmentation. However, this constraint is lost on the CAMs trained by image-level supervision. Therefore, we propose consistency regularization on predicted CAMs from various transformed images to provide self-supervision for network learning. Moreover, we propose a pixel correlation module (PCM), which exploits context appearance information and refines the prediction of current pixel by its similar neighbors, leading to further improvement on CAMs consistency. Extensive experiments on PASCAL VOC 2012 dataset demonstrate our method outperforms state-of-the-art methods using the same level of supervision. The code is released online\footnote{https://github.com/YudeWang/SEAM}.
	\end{abstract}
	
	\section{Introduction}
	
	Semantic segmentation is a fundamental computer vision task, which aims to predict pixel-wise classification results on images. Thanks to the booming of deep learning researches in recent years, the performance of semantic segmentation model has achieved great progress~\cite{DeepLabv2, FCN, PSPNet}, promoting many practical applications, \eg, autopilot and medical image analysis. However, compared to other tasks such as classification and detection, semantic segmentation needs to collect pixel-level class labels which are time-consuming and expensive. Recently many efforts are devoted to weakly supervised semantic segmentation (WSSS) which utilizes weak supervisions, \eg, image-level classification labels, scribbles, and bounding boxes, attempting to achieve equivalent segmentation performance of fully supervised approaches. This paper focuses on semantic segmentation by image-level classification labels.
	
	\begin{figure}[t]
		\centering
		\includegraphics[width=1.0\columnwidth]{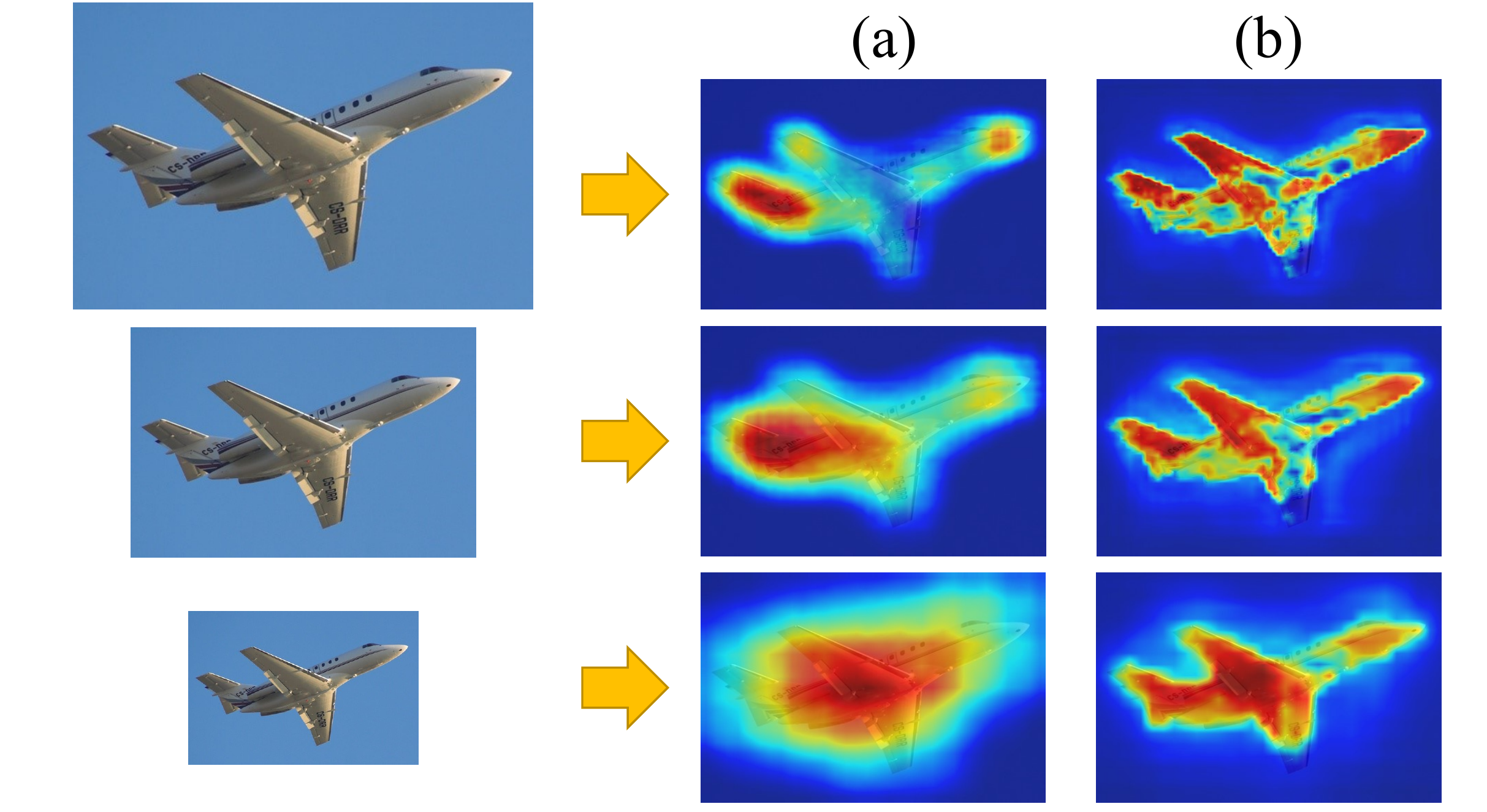}
		\caption{Comparisons of CAMs generated by input images with different scales. (a) Conventional CAMs. (b) CAMs predicted by our SEAM, which are more consistent over rescaling.}
		\label{fig:illustration}
	\end{figure}
	
	To the best of our knowledge, most of advanced WSSS methods are based on the class activation map (CAM)~\cite{CAM}, which is an effective way to localize objects by image classification labels. However, the CAMs usually only cover the most discriminative part of the object and incorrectly activate in background regions, which can be summarized as under-activation and over-activation respectively. Moreover, the generated CAMs are not consistent when images are augmented by affine transformations. As shown in Fig.~\ref{fig:illustration}, applying different rescaling transformations on the same input images causes significant inconsistency on the generated CAMs. The essential causes of these phenomena come from the supervision gap between fully and weakly supervised semantic segmentation.
	
    In this paper, we propose a self-supervised equivariant attention mechanism (SEAM) to narrow the supervision gap mentioned above. The SEAM applies consistency regularization on CAMs from various transformed images to provide self-supervision for network learning. To further improve the network prediction consistency, SEAM introduces the pixel correlation module (PCM), which captures context appearance information for each pixel and revises original CAMs by learned affinity attention maps. The SEAM is implemented by a siamese network with equivariant cross regularization (ECR) loss, which regularizes the original CAMs and the revised CAMs on different branches. Fig.~\ref{fig:illustration} shows that our CAMs are consistent over various transformed input images, with fewer over-activated and under-activated regions than baseline. Extensive experiments give both quantitative and qualitative results, demonstrating the superiority of our approach.
	
	In summary, our main contributions:
	\begin{itemize}
		\item We propose a self-supervised equivariant attention mechanism (SEAM), incorporating equivariant regularization with pixel correlation module (PCM), to narrow the supervision gap between fully and weakly supervised semantic segmentation.    
		\item The design of siamese network architecture with equivariant cross regularization (ECR) loss efficiently couples the PCM and self-supervision, producing CAMs with both fewer over-activated and under-activated regions.
		\item Experiments on PASCAL VOC 2012 illustrate that our algorithm achieves state-of-the-art performance with only image-level annotations. 
	\end{itemize}
	
	\section{Related Work}
	
	The development of deep learning has led to a series of breakthroughs on fully supervised semantic segmentation~\cite{DeepLabv2, DANet, FCN, OCNet, PSPNet} in recent years. In this section, we introduce some works, including weakly supervised semantic segmentation and self-supervised learning. 
	
	\subsection{Weakly Supervised Semantic Segmentation}
	
	Compared to fully supervised learning, WSSS uses weak labels to guide network training, \eg, bounding boxes~\cite{boxsup, SDI}, scribbles~\cite{scribblesup, RAWK} and image-level classification labels~\cite{SEC, CCNN, MIL}. A group of advanced researches utilizes image-level classification labels to train models. Most of them refine the class activation map (CAM)~\cite{CAM} generated by the classification network to approximate the segmentation mask. SEC~\cite{SEC} proposes three principles, \textit{i}.\textit{e}., seed, expand, and constrain, to refine CAMs, which are followed by many other works. Adversarial erasing~\cite{SeeNet, AdvErasing} is a popular CAM expansion method, which erases the most discriminative part of CAM, guides the network to learn classification features from other regions and expands activations. AffinityNet~\cite{AffinityNet} trains another network to learn the similarity between pixels, which generates a transition matrix and multiplies with CAM several times to adjust its activation coverage. IRNet~\cite{IRNet} generates a transition matrix from the boundary activation map and extends the method to weakly supervised instance segmentation. Here are also some researches endeavor to aggregate self-attention module~\cite{selfattention, nonlocal} in the WSSS framework, \eg, CIAN~\cite{CIAN} proposes cross-image attention module to learn activation maps from two different images containing the same class objects with the guidance of saliency maps.
	
	\subsection{Self-supervised Learning}
	
	Instead of using massive annotated labels to train network, self-supervised learning approaches aim at designing pretext tasks to generate labels without additional manual annotations. Here are many classical self-supervised pretext tasks, \eg, relative position prediction~\cite{doersch2015unsupervised}, spatial transformation prediction~\cite{gidaris2018unsupervised}, image inpainting~\cite{pathak2016context}, and image colorization~\cite{larsson2016learning}. To some extent, the generative adversarial network~\cite{GAN} can also be regarded as a self-supervised learning approach that the authenticity labels for discriminator do not need to be annotated manually. Labels generated by pretext tasks provide self-supervision for the network to learn a more robust representation. The feature learned by self-supervision can replace the feature pretrained by ImageNet~\cite{imagenet} on some tasks, such as detection~\cite{doersch2015unsupervised} and part segmentation~\cite{SCOPS}. 
	
	Considering there is a large supervision gap between fully and weakly supervised semantic segmentation, it is an intuition that we should seek additional supervision to narrow the gap. Since image-level classification labels are too weak for network to learn segmentation masks which should well fit object boundary, we design pretext task using the equivariance of ideal segmentation function to provide additional self-supervision for network learning with only image-level annotations.

	\section{Approach}
	\begin{figure*}[t]
		\centering
		\includegraphics[width=0.9\linewidth]{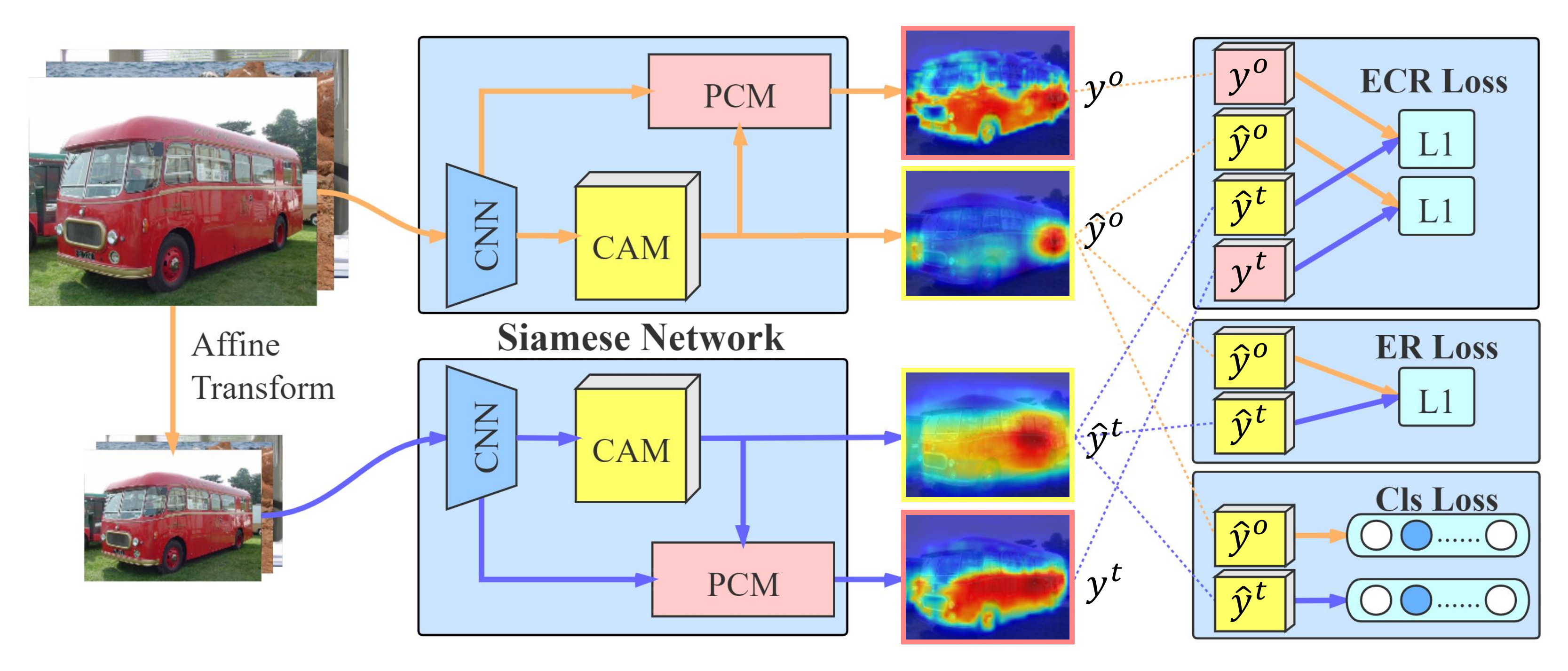}
		\caption{The siamese network architecture of our proposed SEAM method. The SEAM is the integration of equivariant regularization (ER) (Section.~\ref{subsec:er}) and pixel correlation module (PCM) (Section.~\ref{subsec:PCM}). With specially designed losses (Section~\ref{subsec:loss}), the revised CAMs not only keep consistent over affine transformation but also well fit the object contour.}
		\label{fig:network}
	\end{figure*}
	This section details our SEAM method. Firstly, we illustrate the motivation of our work. Then we introduce the implementation of equivariant regularization by a shared-weight siamese network. The proposed pixel correlation module (PCM) is integrated into the network to further improve the consistency of prediction. Finally, the loss design of SEAM is discussed. Fig.~\ref{fig:network} shows our SEAM network structure.
	
	\subsection{Motivation}
	We denote ideal pixel-level semantic segmentation function as $F_{w_s}(\cdot)$ with parameters $w_s$. For each image sample $I$, the segmentation process can be formulated as $F_{w_s}(I)=s$, where $s$ denotes pixel-level segmentation mask. The formulation is also consistent in classification task. With additional image-level label $l$ and pooling function $\mathrm{Pool}(\cdot)$, classification task can be represented as $\mathrm{Pool}(F_{w_c}(I))=l$ with parameters $w_c$. Most WSSS approaches are based on the hypothesis that the optimal parameters for classification and segmentation satisfy $w_c=w_s$. Therefore, these methods train a classification network firstly and remove pooling function to tackle segmentation task.
	
	However, it is easy to find the properties of classification and segmentation function are different. Suppose there is an affine transformation $A(\cdot)$ for each sample, the segmentation function is more inclined to be equivariant, \textit{i}.\textit{e}., $F_{w_s}(A(I))=A(F_{w_s}(I))$. While the classification task focuses more on invariance, \textit{i}.\textit{e}., $\mathrm{Pool}(F_{w_c}(A(I)))=l$. Although the invariance of classification function is mainly caused by pooling operation, there is no equivariant constraint for $F_{w_c}(\cdot)$, which makes it nearly impossible to achieve the same objective of segmentation function during network learning. Additional regularizers should be integrated to narrow the supervision gap between fully and weakly supervised learning.
	
	Self-attention is a widely accepted mechanism that can significantly improve the network approximation ability. It revises feature maps by capturing context feature dependency, which also meets the ideas of most WSSS methods using the similarity of pixels to refine the original activation map. Following the denotation of~\cite{nonlocal}, the general self-attention mechanism can be defined as:
	\begin{equation}
	\label{eq:nonlocal}
	\mathrm{y}_i = \frac{1}{\mathcal{C}(\mathrm{x}_i)}\sum_{\forall j}f(\mathrm{x}_i, \mathrm{x}_j)g(\mathrm{x}_j)+\mathrm{x}_i,
	\end{equation}
	\begin{equation}
	\label{eq:gaussianfunc}
	f(\mathrm{x}_i, \mathrm{x}_j)=e^{\theta(\mathrm{x}_i)^\mathrm{T}\phi(\mathrm{x}_j)}.
	\end{equation}
	Here $\mathrm{x}$ and $\mathrm{y}$ denote input and output feature, with spatial position index $i$ and $j$. The output signal is normalized by $\mathcal{C}(\mathrm{x}_i)=\sum_{\forall j}f(\mathrm{x}_i,\mathrm{x}_j)$. Function $g(\mathrm{x}_j)$ gives a representation of input signal $\mathrm{x}_j$ at each position and all of them are aggregated into position $i$ with the similarity weights given by $f(\mathrm{x}_i,\mathrm{x}_j)$, which calculates the dot-product pixel affinity in an embedding space. To improve the network ability for consistent prediction, we propose SEAM by incorporating self-attention with equivariant regularization.
	
	\subsection{Equivariant Regularization}
	\label{subsec:er}
	
	During the data augmentation period of fully supervised semantic segmentation, the pixel-level labels should be applied with the same affine transformation as input images. It introduces an implicit equivariant constraint for the network. However, considering that the WSSS can only access image-level classification labels, the implicit constraint is missing here. Therefore, we propose equivariant regularization as follows:
	\begin{equation}
	\label{eq:er}
	\mathcal{R}_{ER}=||F(A(I))-A(F(I))||_1.
	\end{equation}
	Here $F(\cdot)$ denotes the network, and $A(\cdot)$ denotes any spatial affine transformation, \eg, rescaling, rotation, flip. To integrate regularization on the original network, we expand the network into a shared-weight siamese structure. One branch applies the transformation on the network output, the other branch warps the images by the same transformation before the feedforward of the network. The output activation maps from two branches are regularized to guarantee the consistency of CAMs.
	
	\subsection{Pixel Correlation Module}
	\label{subsec:PCM}
	Although equivariant regularization provides additional supervision for network learning, it is hard to achieve ideal equivariance with only classical convolution layers. Self-attention is an efficient module to capture context information and refine pixel-wise prediction results. To integrate the classical self-attention module given by Eq.~(\ref{eq:nonlocal}) and Eq.~(\ref{eq:gaussianfunc}) for CAM refinement, the formulation can be written as:
	\begin{equation}
	\label{eq:clsselfattention}
	\mathrm{y}_i=\frac{1}{\mathcal{C}(\mathrm{x}_i)}\sum_{\forall j}e^{\theta(\mathrm{x}_i)^\mathrm{T}\phi(\mathrm{x}_j)}g(\mathrm{\hat{y}}_j)+\mathrm{\hat{y}}_i,
	\end{equation}
	where $\mathrm{\hat{y}}$ denotes the original CAM and $\mathrm{y}$ denotes the revised CAM. In this structure, the original CAM is embedded into residual space by function $g$. Each pixel aggregates with others with similarity given by Eq.~(\ref{eq:gaussianfunc}). Three embedding functions $\theta, \phi, g$ can be implemented by individual $1\times 1$ convolution layers.
	
\begin{figure}[t]
		\centering
		\includegraphics[width=1.0\columnwidth]{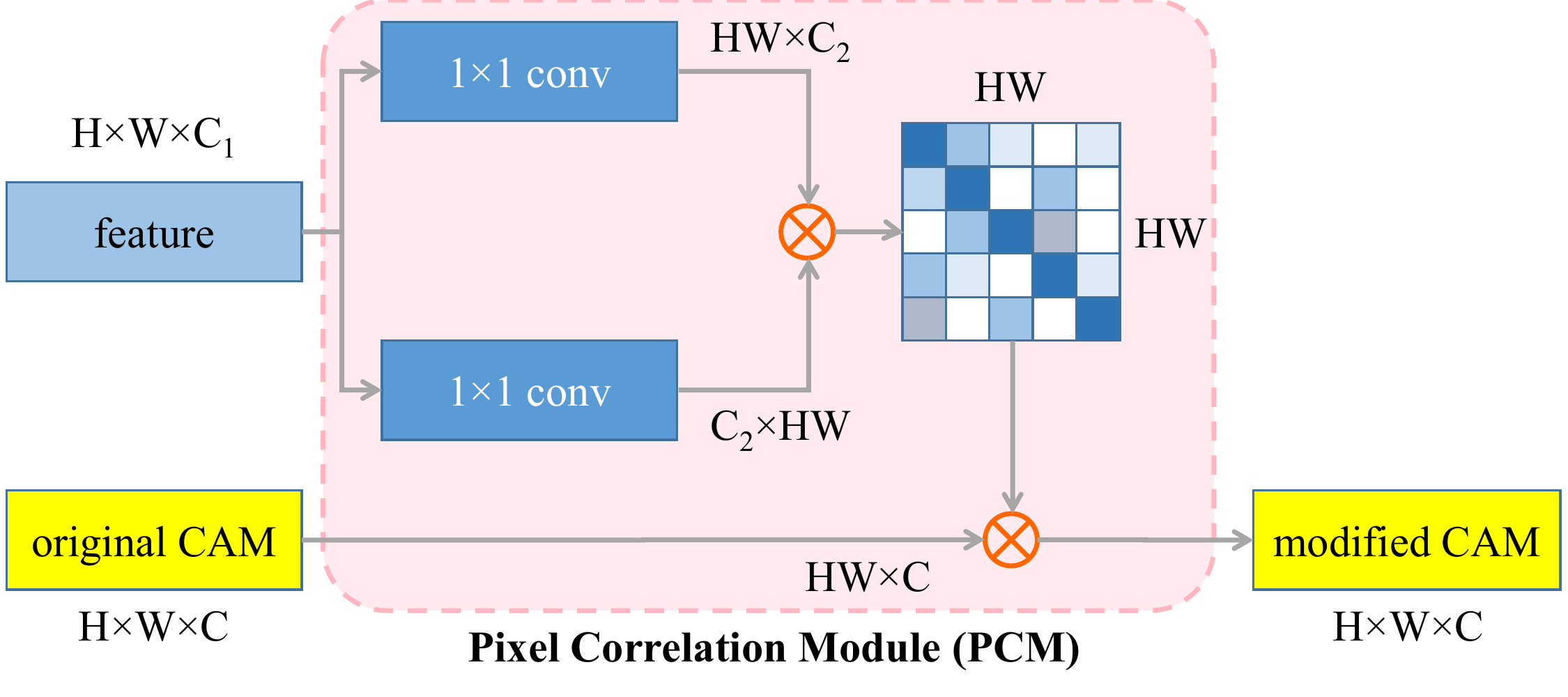}
		\caption{The structure of PCM, where $H, W, C/C_1/C_2$ denote height, width and channel numbers of feature maps respectively.}
		\label{fig:PCM}
	\end{figure}

	To further refine original CAMs by context information, we propose a pixel correlation module (PCM) at the end of the network to integrate the low-level feature of each pixel. The structure of PCM refers to the core part of the self-attention mechanism with some modifications and trained by the supervision from equivariant regularization. We use cosine distance to evaluate inter-pixel feature similarity:
	\begin{equation}
	\label{eq:cos}
	f(\mathrm{x}_i, \mathrm{x}_j)=\frac{\theta(\mathrm{x}_i)^\mathrm{T}\theta(\mathrm{x}_j)}{||\theta(\mathrm{x}_i)||\cdot||\theta(\mathrm{x}_j)||}.
	\end{equation}
	Here we take the inner-product in normalized feature space to calculate the affinity between current pixel $i$ and others. The $f$ can be integrated into Eq.~(\ref{eq:nonlocal}) with some modifications as:
	\begin{equation}
	\label{eq:structure3}
	\mathrm{y}_i=\frac{1}{\mathcal{C}(\mathrm{x}_i)}\sum_{\forall j}\mathrm{ReLU}(\frac{\theta(\mathrm{x}_i)^\mathrm{T}\theta(\mathrm{x}_j)}{||\theta(\mathrm{x}_i)||\cdot ||\theta(\mathrm{x}_j)||})\mathrm{\hat{y}}_j.
	\end{equation}
	The similarities are activated by ReLU to suppress negative values. The final CAM is the weighted sum of the original CAM with normalized similarities. Fig.~\ref{fig:PCM} gives an illustration of the PCM structure.
	
	Compared to classical self-attention, PCM removes the residual connection to keep the same activation intensity of the original CAM. Moreover, since the other network branch provides pixel-level supervision for PCM, which is not as accurate as ground truth, we reduce parameters by removing embedding function $\phi$ and $g$ to avoid overfitting on inaccurate supervision. We use ReLU activation function with L1 normalization to mask out irrelevant pixels and generate an affinity attention map which is smoother in relevant regions. 
	
	\subsection{Loss Design of SEAM}
	\label{subsec:loss}
	Image-level classification label $l$ is the only human-annotated supervision that can be used here. We employ the global average pooling layer at the end of the network to achieve prediction vector $\mathrm{z}$ for image classification and adopt multi-label soft margin loss for network training. The classification loss is defined for $C-1$ foreground object category as:
	\begin{equation}
	\begin{aligned}
	\ell_{cls}(\mathrm{z},l)=-\frac{1}{C-1}\sum_{c=1}^{C-1} & [l_c\log(\frac{1}{1+e^{-\mathrm{z}_c}})\\
	& +(1-l_c)\log(\frac{e^{-\mathrm{z}_c}}{1+e^{-\mathrm{z}_c}})].
	\end{aligned}
	\end{equation}
	Formally we denote the original CAMs of siamese network as $\hat{\mathrm{y}}^o$ and $\hat{\mathrm{y}}^t$, where $\hat{\mathrm{y}}^o$ comes from the branch with original image input and $\hat{\mathrm{y}}^t$ stems from the transformed images. The global average pooling layer aggregates them into prediction vector $\mathrm{z}^o$ and $\mathrm{z}^t$ respectively. The classification loss is calculated on two branches as:
	\begin{equation}
	\mathcal{L}_{cls}=\frac{1}{2}(\ell_{cls}(\mathrm{z}^o,l)+\ell_{cls} (\mathrm{z}^t,l)).
	\end{equation}	
	The classification loss provides learning supervision for object localization. And it is necessary to aggregate equivariant regularization on original CAM to preserve the consistency of output. The equivariant regularization (ER) loss on original CAM can be easily defined as:
	\begin{equation}
	\mathcal{L}_{ER} = ||A(\hat{\mathrm{y}}^o)-\hat{\mathrm{y}}^t||_1.
	\end{equation}
	Here $A(\cdot)$ is an affine transformation which has already been applied to the input image in the transformation branch of the siamese network. Moreover, to further improve the ability of network for equivariance learning, the original CAMs and features from the shallow layers are fed into PCM for refinement. The intuitive idea is introducing equivariant regularization between revised CAMs $\mathrm{y}^o$ and $\mathrm{y}^t$. However, in our early experiments, the output maps of PCM fall into the local minimum quickly that all pixels in the image are predicted the same class. Therefore, we propose an equivariant cross regularization (ECR) loss as:
	\begin{equation}
	\mathcal{L}_{ECR} = ||A(\mathrm{y}^o)-\hat{\mathrm{y}}^t||_1+||A(\hat{\mathrm{y}}^o)-\mathrm{y}^t||_1.
	\label{eq:ecrloss}
	\end{equation} 
	The PCM outputs are regularized by the original CAMs on the other branch of the siamese network. This strategy can avoid CAM degeneration during PCM refinement.
	
	Although the CAMs are learned by foreground object classification loss, there are many background pixels, which should not be ignored during PCM processing. The original foreground CAMs have zero vectors on these background positions, which cannot produce gradients to push feature representations closer between those background pixels. Therefore, we define the background score as:
	\begin{equation}
	\hat{\mathrm{y}}_{i,bkg}=1-\max_{1\leq c\leq C-1}\hat{\mathrm{y}}_{i,c},
	\end{equation}
	where $\hat{\mathrm{y}}_{i,c}$ is the activation score of original CAM for category $c$ at position $i$. We normalize the activation vectors of each pixel by suppressing foreground non-maximum activations to zeros and concatenate with additional background score. During inference, we only keep the foreground activation results and set the background score as $\hat{\mathrm{y}}_{i,bkg}=\alpha$, where $\alpha$ is the hard threshold parameter.
	
	In summary, the final loss of SEAM is defined as:
	\begin{equation}
	\mathcal{L} = \mathcal{L}_{cls} + \mathcal{L}_{ER} + \mathcal{L}_{\mathit{ECR}}.
	\end{equation}
	The classification loss is used to roughly localize objects and the ER loss is used to narrow the gaps between pixel- and image-level supervisions. The ECR loss is used to integrate PCM with the trunk of the network, in order to make consistent predictions over various affine transformations. The network architecture is illustrated in Fig.~\ref{fig:network}. We give the details of network training settings and carefully investigate the effectiveness of each module in the experiments section.
	
	\section{Experiments}
	
	\subsection{Implementation Details}
    We evaluate our approach on PASCAL VOC 2012 dataset with 21 class annotations, \textit{i}.\textit{e}., 20 foreground objects and the background. The official dataset separation has 1464 images for training, 1449 for validation and 1456 for testing. Following the common experimental protocol for semantic segmentation, we take additional annotations from SBD~\cite{SBD} to build an augmented training set with 10582 images. Noting that only image-level classification labels are available during network training. Mean intersection over union (mIoU) is used as a metric to evaluate segmentation results.

	In our experiments, ResNet38~\cite{resnet38} is adopted as backbone network with $output\_stride=8$. We extract the feature maps from stage 3 and stage 4, reduce their channel numbers into 64 and 128 respectively by individual $1\times 1$ convolution layers. In PCM, these features are concatenated with images and fed into function $\theta$ in Eq.~(\ref{eq:cos}), which is implemented by another $1\times 1$ convolution layer. The images are randomly rescaled in the range of [448, 768] by the longest edge and then cropped by $448\times 448$ as network inputs. The model is trained on 4 TITAN-Xp GPUs with batch size 8 for 8 epochs. The initial learning rate is set as 0.01, following the poly policy $\mathit{lr}_{\mathit{itr}}=\mathit{lr}_{\mathit{init}}(1-\frac{itr}{\mathit{max\_itr}})^\gamma$ with $\gamma=0.9$ for decay. Online hard example mining (OHEM) is employed on the ECR loss remaining the largest $20\%$ pixel losses.

	During network training, we cut off gradients back-propagation at the intersection point between PCM stream and the trunk of the network to avoid the mutual interference. This setting simplifies the PCM into a pure context refinement module which still can be trained with the backbone of the network at the same time. And the learning of original CAMs will not be affected by PCM refinement process. During inference, since our SEAM is a shared-weight siamese network, only one branch needs to be restored. We adopt multi-scale and flip test during inference to generate pseudo segmentation labels.
	
	\subsection{Ablation Studies}
	To verify the effectiveness of our SEAM, we generate pixel-level pseudo labels from revised CAMs on PASCAL VOC 2012 \textit{train} set. In our experiments, we traverse all background threshold options and give the best mIoU of pseudo labels, instead of comparing with the same background threshold. Because the highest pseudo label accuracy represents the best matching results between CAMs and ground truth segmentation masks. Specifically, the foreground activation coverage will expand with the increase of average activation intensity, while its matching degree with ground truth is not changed. And the highest pseudo label accuracy will not be improved when CAMs only increase average activation intensity rather than becoming more matchable with ground truth.
	\paragraph{Comparison with Baseline:} 
	
\begin{figure}[tbp]
		\centering
		\includegraphics[width=1.0\columnwidth]{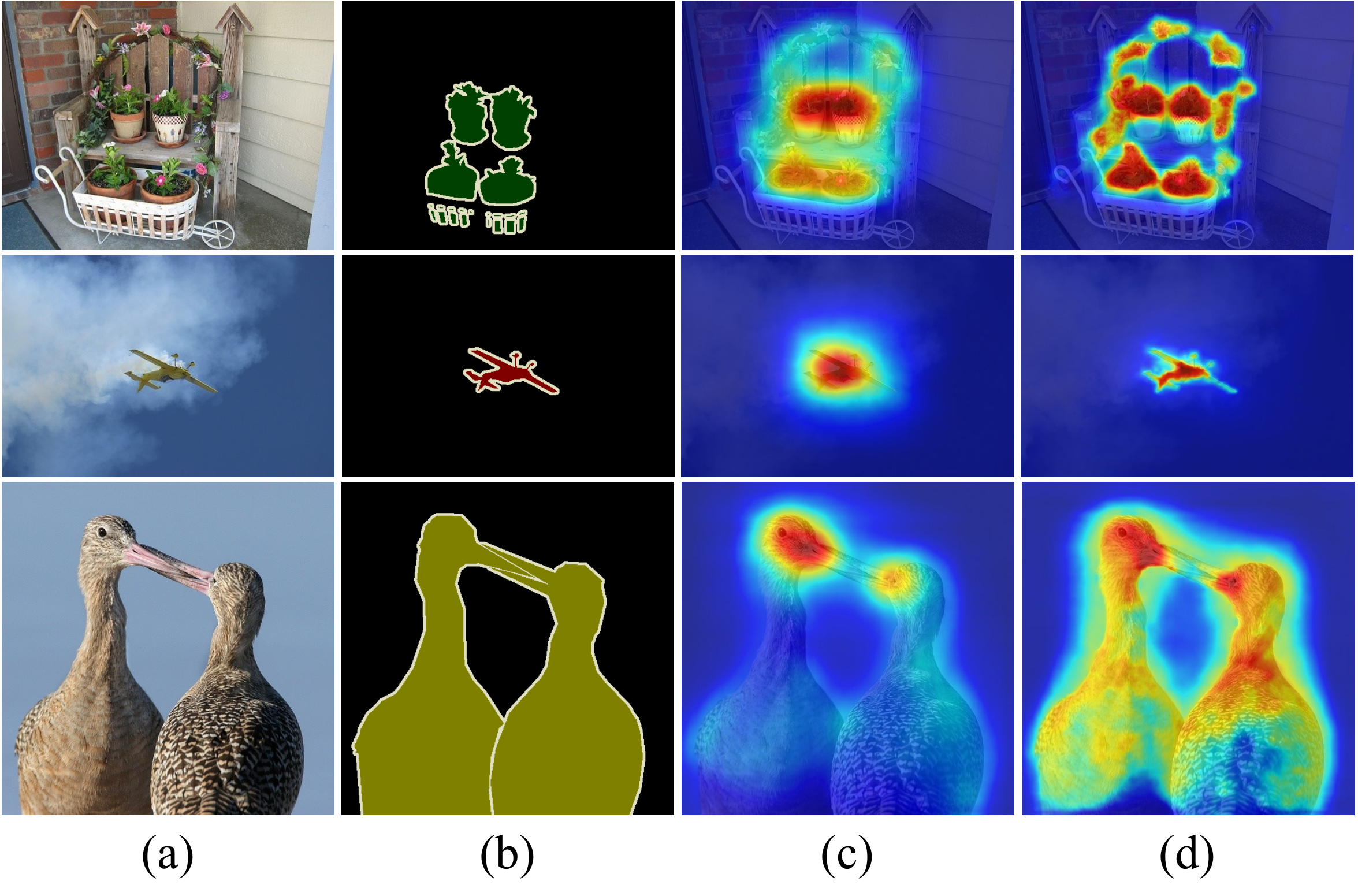}
		\caption{The visualization of CAMs. (a) Original images. (b) Ground truth segmentations. (c) Baseline CAMs. (d) CAMs produced by SEAM. The SEAM not only suppresses over-activation but also expands CAMs into complete object activation coverage.}
		\label{fig:baseline}
	\end{figure}

\begin{figure}[tbp]
		\centering
		\includegraphics[width=1.0\columnwidth]{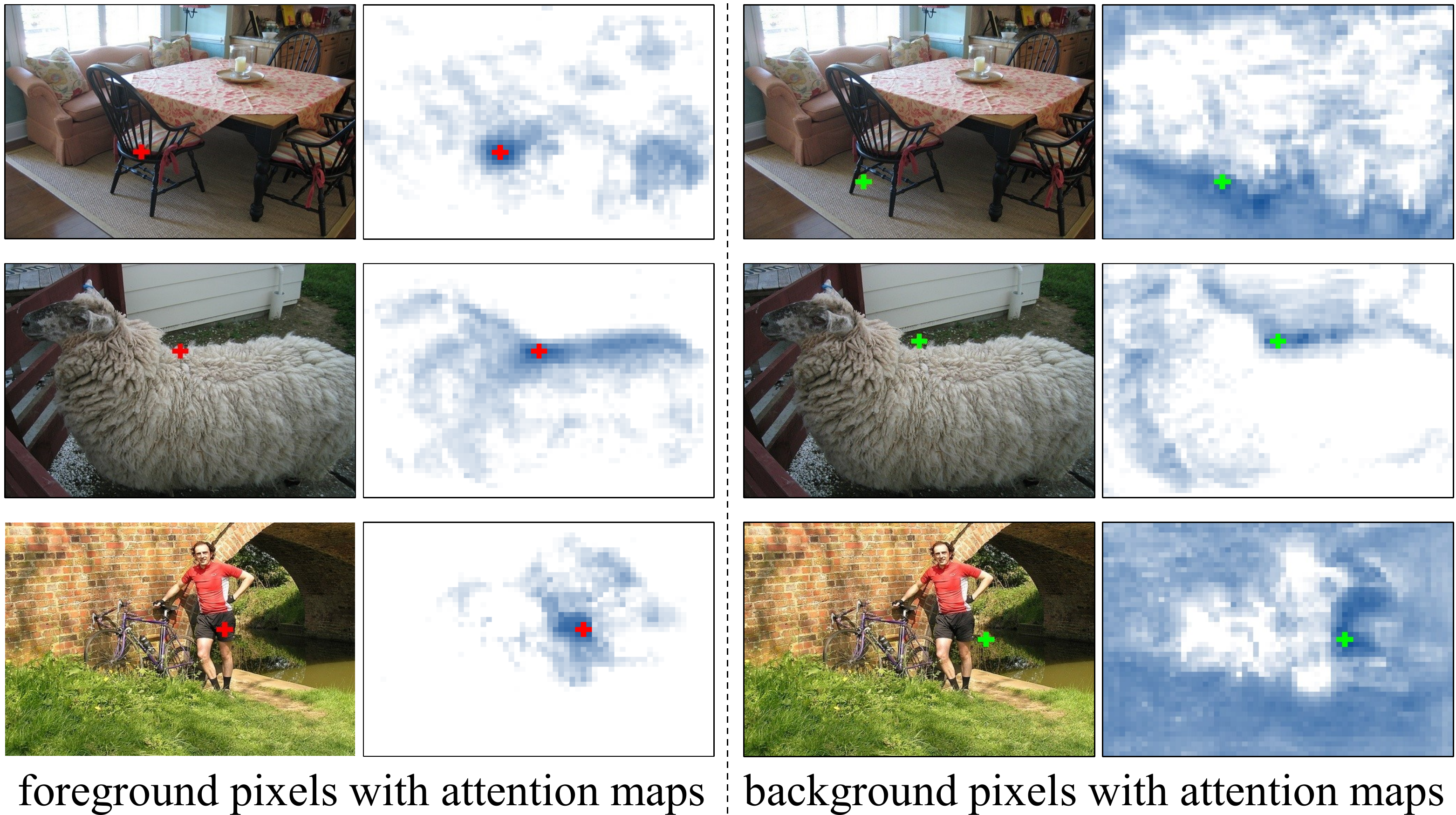}
		\caption{The visualization of affinity attention map on foreground and background. The red and green crosses denote the selected pixels, with similar feature representation in blue color.}
		\label{fig:attmap}
	\end{figure}

	Tab.~\ref{tab:steps} gives an ablation study of each module in our approach. It shows that using the siamese network with equivariant regularization has a 2.47\% improvement compared to baseline. Our PCM achieves significant performance elevation by 5.18\%. After applying OHEM on equivariant cross regularization loss, the generated pseudo labels further achieve 55.41\% mIoU on PASCAL VOC train set. We also test the baseline CAM with dense CRF to refine predictions. The results show that dense CRF improves the mIoU to 52.40\%, which is lower than the SEAM result 55.41\%. And our SEAM can further improve the performance up to 56.83\% after aggregating dense CRF as post process. Fig.~\ref{fig:baseline} shows that the CAMs generated by SEAM have fewer over-activations and more complete activation coverage, whose shape is closer to the ground truth segmentation masks than baseline. To further verify the effectiveness of our proposed SEAM, we visualize the affinity attention maps generated by PCM. As shown in Fig.~\ref{fig:attmap}, the selected foreground and background pixels are very close in spatial, while their affinity attention maps are greatly different. It proves that the PCM can learn boundary sensitive features from self-supervision.
	\begin{table}[tbp]
		\centering
		\begin{tabular}{cccccc}
			\hline
			baseline & ER & PCM & OHEM & CRF & mIoU\\
			\hline
			$\surd$ & & & & & 47.43\%\\		
			$\surd$ & & & & $\surd$ & 52.40\%\\
			$\surd$ & $\surd$ & & & & 49.90\%\\
			$\surd$ & $\surd$ & $\surd$ & & & 55.08\%\\
			$\surd$ & $\surd$ & $\surd$ & $\surd$ & & 55.41\%\\
			$\surd$ & $\surd$ & $\surd$ & $\surd$ & $\surd$ & 56.83\%\\
			\hline
		\end{tabular}
		\caption{The ablation study for each part of SEAM. \textbf{ER}: equivariant regularization. \textbf{PCM}: pixel correlation module. \textbf{OHEM}: online hard example mining. \textbf{CRF}: conditional random field.}
		\label{tab:steps}
	\end{table}
	\begin{table}[t]
		\centering
		\setlength{\tabcolsep}{0.6cm}{
			\begin{tabular}{lc}
				\hline
				model &  mIoU\\
				\hline
				CAM & 47.43\%\\
				GradCAM & 46.53\%\\
				GradCAM++ & 47.37\%\\
				CAM + SEAM & 55.41\%\\
				\hline
			\end{tabular}
		}
		\caption{Evaluation of various weakly supervised localization methods with semantic segmentation metric (mIoU).}
		\label{tab:gradcam}
	\end{table}
	
	\paragraph{Improved Localization Mechanism:}
	
	It is an intuition that improved weakly supervised localization mechanism will elevate mIoU of pseudo segmentation labels. To verify the idea, we simply evaluate GradCAM~\cite{GradCAM} and GradCAM++\cite{GradCAM++} before aggregating our proposed SEAM. However, the evaluation results given by Tab.~\ref{tab:gradcam} illustrates that both GradCAM and GradCAM++ cannot narrow the supervision gap between fully and weakly supervised semantic segmentation tasks, since the best mIoU results do not have improvement. We believe the improved localization mechanisms are only designed to represent object correlated parts without any constraints by low-level information, which is not suitable for the segmentation task. The CAMs generated by these improved localization methods are not becoming more matchable with ground truth masks. The following experiments further illustrate that our proposed SEAM can substantially improve the quality of CAM to fit the shape of object masks. 
	\paragraph{Affine Transformation:}
	Ideally, the $A(\cdot)$ in Eq.~(\ref{eq:er}) can be any affine transformation. Several transformations are conducted in the siamese network to evaluate the effect of them on equivariant regularization. As shown in Tab.~\ref{tab:transformation}, there are four candidate affine transformations: rescaling with 0.3 down-sampling rate, random rotation in [-20, 20] degrees, translation by 15 pixels and horizontal flip. Firstly, our proposed SEAM simply adopts rescaling during network training. Tab.~\ref{tab:transformation} shows that the mIoU of pseudo labels has significant improvement from 47.43\% to 55.41\%. Tab.~\ref{tab:transformation} also shows that simply incorporating different transformations is not much effective. When rescaling transformation integrates with flip, rotation, and translation respectively, only flip makes tiny improvement. In our view, it is because the activation maps between flip, rotation, and translation are too similar to produce sufficient supervision. Without additional instructions, we only preserve rescaling as the key transformation with $0.3$ down-sampling rate in our other experiments.
	\begin{table}[tbp]
		\centering
		\begin{tabular}{ccccc}
			\hline
			rescale & flip & rotation & translation & mIoU\\
			\hline
			& & & & 47.43\%\\
			$\surd$ & & & & 55.41\%\\
			$\surd$ & $\surd$ & & & 55.50\%\\
			$\surd$ & & $\surd$ & & 53.13\%\\
			$\surd$ & & & $\surd$ & 55.23\%\\
			\hline
		\end{tabular}
		\caption{Experiments of various transformations on equivariant regularization. Simply aggregating different affine transformations cannot bring significant improvement.}
		\label{tab:transformation}
	\end{table}
	\begin{table}[tbp]
		\centering
		\begin{tabular}{lcc}
			\hline
			model & random rescale & mIoU\\
			\hline
			baseline & [448, 768] & 47.43\%\\
			baseline & [224, 768] & 46.72\%\\
			SEAM & [448, 768] & 53.47\%\\
			\hline
		\end{tabular}
		\caption{Experiments of augmentation rescaling range. Here the rescale rate of SEAM is set to 0.5.}
		\label{tab:range}
	\end{table}
	\begin{table}[tbp]
		\centering
		\begin{tabular}{l|c|c}
			\hline
			test scale & baseline (mIoU) & ours (mIoU)\\
			\hline
			$[0.5]$ & 40.17\% & 49.35\%\\
			$[1.0]$ & 46.10\% & 51.57\%\\
			$[1.5]$ & 47.51\% & 52.25\%\\
			$[2.0]$ & 46.12\% & 49.79\%\\
			\hline
			$[0.5, 1.0, 1.5, 2.0]$ & 47.43\% & 55.41\%\\
			\hline
		\end{tabular}
		\caption{Experiments with various single- and multi-scale test.}
		\label{tab:singlemultiscale}
	\end{table}
	\paragraph{Augmentation and Inference:}
	Compared to the original one-branch network, the siamese structure expands the augmentation range of image size in practice. To investigate whether the improvement stems from the rescaling range, we evaluate the baseline model with a larger scale range and Tab.~\ref{tab:range} gives the experiment results. It shows that simply increasing the rescaling range cannot improve the accuracy of generated pseudo labels, which proves that the performance improvement comes from the combination of PCM and equivariant regularization instead of data augmentation.
	
	During inference, it is a common practice to employ multi-scale test by aggregating the prediction results from images with different scales to boost the final performance. It can also be regarded as a method to improve the equivariance of predictions. To verify the effectiveness of our propose SEAM, we evaluate the CAMs generated by both single-scale and multi-scale test. Tab.~\ref{tab:singlemultiscale} illustrates that our proposed model outperforms baseline with higher peak performance in both single- and multi-scale test.
	\begin{figure}[t]
		\centering
		\includegraphics[width=0.8\columnwidth]{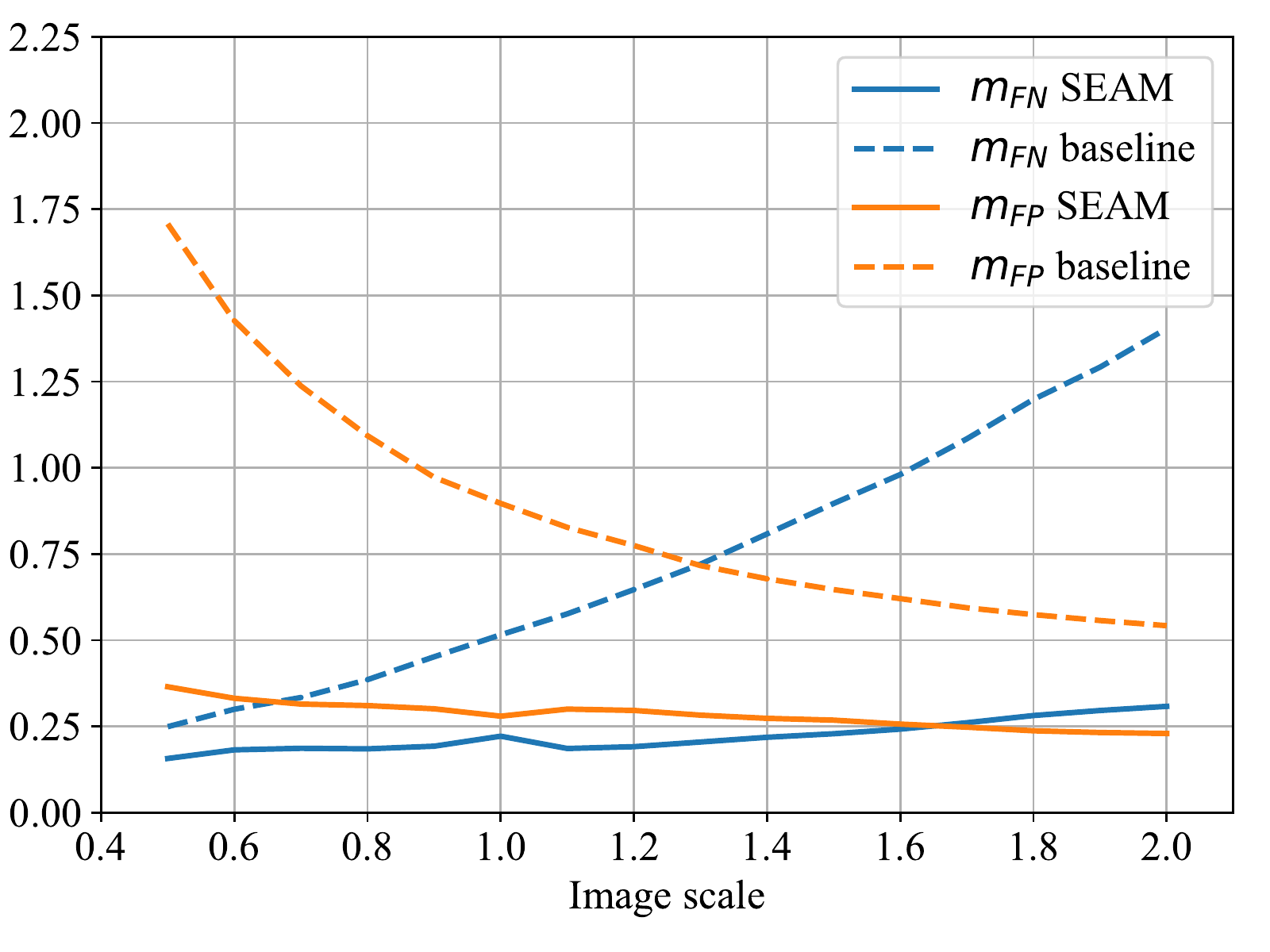}
		\caption{The curves of over-activation and under-activation. Lower $m_{\mathit{FN}}$ curve represents fewer under-activation regions, and lower $m_{\mathit{FP}}$ represents fewer over-activated regions.}
		\label{fig:source}
	\end{figure}
	\begin{figure*}[t]
		\centering
		\includegraphics[width=1.0\linewidth]{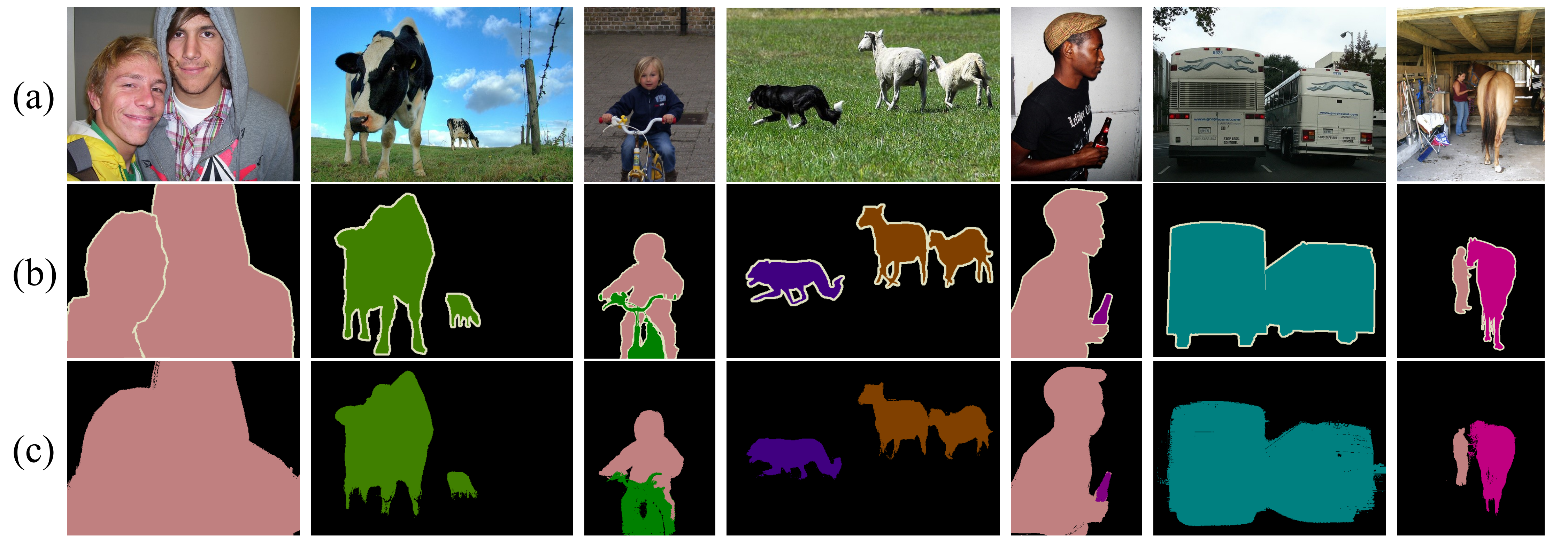}
		\caption{Qualitative segmentation results on PASCAL VOC 2012 \textit{val} set. (a) Original images. (b) Ground truth. (c) Segmentation results predicted by DeepLab model retrained on our pseudo labels.}
		\label{fig:deeplab}
	\end{figure*}
	\begin{table*}[t]
		\centering
		\small
		\setlength{\tabcolsep}{0.4mm}{
			\begin{tabular}{l|ccccccccccccccccccccc|c}
				\hline
				model & bkg & aero & bike & bird & boat & bottle & bus & car & cat & chair & cow & table & dog & horse & mbk & person & plant & sheep & sofa & train & tv & mIoU\\
				\hline
				CCNN~\cite{CCNN} & 68.5 & 25.5 & 18.0 & 25.4 & 20.2 & 36.3 & 46.8 & 47.1 & 48.0 & 15.8 & 37.9 & 21.0 & 44.5 & 34.5 & 46.2 & 40.7 & 30.4 & 36.3 & 22.2 & 38.8 & 36.9 & 35.3\\
				MIL+seg~\cite{MIL} & 79.6 & 50.2 & 21.6 & 40.9 & 34.9 & 40.5 & 45.9 & 51.5 & 60.6 & 12.6 & 51.2 & 11.6 & 56.8 & 52.9 & 44.8 & 42.7 & 31.2 & 55.4 & 21.5 & 38.8 & 36.9 & 42.0\\
				SEC~\cite{SEC} & 82.4 & 62.9 & 26.4 & 61.6 & 27.6 & 38.1 & 66.6 & 62.7 & 75.2 & 22.1 & 53.5 & 28.3 & 65.8 & 57.8 & 62.3 & 52.5 & 32.5 & 62.6 & 32.1 & 45.4 & 45.3 & 50.7\\
				AdvErasing~\cite{AdvErasing} & 83.4 & 71.1 & 30.5 & 72.9 & 41.6 & 55.9 & 63.1 & 60.2 & 74.0 & 18.0 & 66.5 & 32.4 & 71.7 & 56.3 & 64.8 & 52.4 & 37.4 & 69.1 & 31.4 & 58.9 & 43.9 & 55.0\\		
				AffinityNet~\cite{AffinityNet} & 88.2 & 68.2 & 30.6 & 81.1 & \textbf{49.6} & 61.0 & 77.8 & 66.1 & 75.1 & 29.0 & 66.0 & 40.2 & \textbf{80.4} & 62.0 & 70.4 & 73.7 & 42.5 & 70.7 & \textbf{42.6} & \textbf{68.1} & 51.6 & 61.7\\
				\hline
				\textbf{Our SEAM} & \textbf{88.8} & \textbf{68.5} & \textbf{33.3} & \textbf{85.7} & 40.4 & \textbf{67.3} & \textbf{78.9} & \textbf{76.3} & \textbf{81.9} & \textbf{29.1} & \textbf{75.5} & \textbf{48.1} & 79.9 & \textbf{73.8} & \textbf{71.4} & \textbf{75.2} & \textbf{48.9} & \textbf{79.8} & 40.9 & 58.2 & \textbf{53.0} & \textbf{64.5}\\
				\hline
		\end{tabular}}
		\caption{Category performance comparisons on PASCAL VOC 2012 \textit{val} set with only image-level supervision.}
		\label{tab:finalval}
	\end{table*}
	\begin{table}[t]
		\centering
		\small
		\begin{tabular}{lcc|cc}
			\hline
			Methods & Backbone & Saliency & \textit{val} & \textit{test}\\
			\hline
			CCNN~\cite{CCNN} & VGG16 & & 35.3 & 35.6\\
			EM-Adapt~\cite{EM-Adapt} & VGG16 & & 38.2 & 39.6\\
			MIL+seg~\cite{MIL} & OverFeat & & 42.0 & 43.2\\
			SEC~\cite{SEC} & VGG16 & & 50.7 & 51.1\\
			STC~\cite{STC} & VGG16 & $\surd$ &49.8 & 51.2\\
			AdvErasing~\cite{AdvErasing} & VGG16 & $\surd$ & 55.0 & 55.7\\
			MDC~\cite{MDC} & VGG16 & $\surd$ & 60.4 & 60.8\\
			MCOF~\cite{MCOF} & ResNet101 &  $\surd$ & 60.3 & 61.2\\
			DCSP~\cite{DCSP} & ResNet101 & $\surd$ & 60.8 & 61.9\\
			SeeNet~\cite{SeeNet} & ResNet101 & $\surd$ & 63.1 & 62.8\\
			DSRG~\cite{DSRG} & ResNet101 & $\surd$ & 61.4 & 63.2\\
			AffinityNet~\cite{AffinityNet} & ResNet38 & & 61.7 & 63.7\\
			CIAN~\cite{CIAN} & ResNet101 & $\surd$ & 64.1 & 64.7\\
			IRNet~\cite{IRNet} & ResNet50 & & 63.5 & 64.8\\
			FickleNet~\cite{FickleNet} & ResNet101 & $\surd$ & 64.9 & 65.3\\
			\hline
			\textbf{Our baseline} & ResNet38 & & 59.7 & 61.9\\
			\textbf{Our SEAM} & ResNet38 & & 64.5 & 65.7\\
			\hline
		\end{tabular}
		\caption{Performance comparisons of our method with other state-of-the-art WSSS methods on PASCAL VOC 2012 dataset.}
		\label{tab:final}
	\end{table}
	\paragraph{Source of Improvement:}
	The improvement of CAM quality mainly stems from more complete activation coverage or fewer over-activated regions. To further analyze the improvement source of our SEAM, we define two metrics to represent the degree of under-activation and over-activation:
	\begin{equation}\label{eq:mfn}
	m_{\mathit{FN}} =\frac{1}{C-1}\sum_{c=1}^{C-1}\frac{\mathit{FN}_c}{\mathit{TP}_c},
	\end{equation}
	\begin{equation}\label{eq:mfp}
	m_{\mathit{FP}} =\frac{1}{C-1}\sum_{c=1}^{C-1}\frac{\mathit{FP}_c}{\mathit{TP}_c}.
	\end{equation}
	Here $\mathit{TP}_c$ denotes the pixel number of true positive prediction of class $c$, $\mathit{FP}_c$ and $\mathit{FN}_c$ denote false positive and false negative respectively. These two metrics exclude the background category since the prediction of background is inverse to the foreground. Specifically, if there are more false negative regions when CAMs do not have complete activation coverage, $m_{\mathit{FN}}$ will have a larger value. Relatively, larger $m_{\mathit{FP}}$ means there are more false positive regions, meaning that CAMs are over-activated.
	
	Based on these two metrics, we collect the evaluation results from both baseline and our SEAM, then plot the curves in Fig.~\ref{fig:source} which illustrates a large gap between baseline and our method. The SEAM achieves lower $m_{\mathit{FN}}$ and $m_{\mathit{FP}}$, meaning that the CAMs generated by our approach have more complete activation coverage and fewer over-activated pixels. Therefore, the prediction maps of SEAM better fit the shape of ground truth segmentation. Moreover, the curves of SEAM are more consistent than baseline model over different image scales, which proves that the equivariance regularization works during network learning and contributes to the improvement of CAM.
	
	\subsection{Comparison with State-of-the-arts}
	To further elevate the accuracy of pseudo pixel-level annotations, we follow the work of~\cite{AffinityNet} to train an AffinityNet based on our revised CAM. The final synthesized pseudo labels achieve 63.61\% mIoU on PASCAL VOC 2012 train set. Then we train the classical segmentation model DeepLab~\cite{DeepLabv1} with ResNet38 backbone on these pseudo labels in full supervision to achieve final segmentation results. Tab.~\ref{tab:finalval} shows the mIoU of each class on \textit{val} set and Tab.~\ref{tab:final} gives more experiment results of previous approaches. Compared to the baseline method, our SEAM significantly improves the performance on both \textit{val} and \textit{test} set with the same training setting. Moreover, our method presents the state-of-the-art performance using only image-level labels on PASCAL VOC 2012 \textit{test} set. Noting that our performance elevation stems from neither the larger network structure nor the improved saliency detector. The performance improvement mainly comes from the cooperation of additional self-supervision and PCM, which produces better CAMs for the segmentation task. Fig.~\ref{fig:deeplab} shows some qualitative results, which verify that our method works well on both large and small objects. 
	
	\section{Conclusion}
	In this paper, we propose a self-supervised equivariant attention mechanism (SEAM) to narrow the supervision gap between fully and weakly supervised semantic segmentation by introducing additional self-supervision. The SEAM embeds self-supervision into weakly supervised learning framework by exploiting equivariant regularization, which forces CAMs predicted from various transformed images to be consistent. To further improve the ability of network for generating consistent CAMs, a pixel correlation module (PCM) is designed, which refines original CAMs by learning inter-pixel similarity. Our SEAM is implemented by a siamese network structure with efficient regularization losses. The generated CAMs not only keep consistent over different transformed inputs but also better fit the shape of ground truth masks. The segmentation network retrained by our synthesized pixel-level pseudo labels achieves state-of-the-art performance on PASCAL VOC 2012 dataset, which proves the effectiveness of our SEAM.

	\paragraph{Acknowledgement:}
	This work was partially supported by National Key R\&D Program of China (No. 2017YFA0700800), CAS\; Frontier\; Science\; Key\; Research\; Project\; (No.\; QYZDJ-SSWJSC009) and Natural Science Foundation of China (Nos. 61806188, 61772496).
	
	{\small
		\bibliographystyle{ieee_fullname}
		\bibliography{egbib}

\begin{thebibliography}{10}\itemsep=-1pt

\bibitem{IRNet}
Jiwoon Ahn, Sunghyun Cho, and Suha Kwak.
\newblock Weakly supervised learning of instance segmentation with inter-pixel
  relations.
\newblock In {\em Proc. IEEE Conference on Computer Vision and Pattern
  Recognition (CVPR)}, 2019.

\bibitem{AffinityNet}
Jiwoon Ahn and Suha Kwak.
\newblock Learning pixel-level semantic affinity with image-level supervision
  for weakly supervised semantic segmentation.
\newblock In {\em Proc. IEEE Conference on Computer Vision and Pattern
  Recognition (CVPR)}, 2018.

\bibitem{GradCAM++}
Aditya Chattopadhay, Anirban Sarkar, Prantik Howlader, and Vineeth~N
  Balasubramanian.
\newblock Grad-cam++: Generalized gradient-based visual explanations for deep
  convolutional networks.
\newblock 2018.

\bibitem{DCSP}
Arslan Chaudhry, Puneet~K Dokania, and Philip~HS Torr.
\newblock Discovering class-specific pixels for weakly-supervised semantic
  segmentation.
\newblock In {\em Proc. British Machine Vision Conference (BMVC)}, 2017.

\bibitem{DeepLabv1}
Liang-Chieh Chen, George Papandreou, Iasonas Kokkinos, Kevin Murphy, and Alan~L
  Yuille.
\newblock Semantic image segmentation with deep convolutional nets and fully
  connected crfs.
\newblock In {\em Proc. International Conference on Learning Representations
  (ICLR)}, 2015.

\bibitem{DeepLabv2}
Liang-Chieh Chen, George Papandreou, Iasonas Kokkinos, Kevin Murphy, and Alan~L
  Yuille.
\newblock Deeplab: Semantic image segmentation with deep convolutional nets,
  atrous convolution, and fully connected crfs.
\newblock {\em IEEE Transactionson Pattern Analysis and Machine Intelligence
  (TPAMI)}, 40(4):834--848, 2018.

\bibitem{boxsup}
Jifeng Dai, Kaiming He, and Jian Sun.
\newblock Boxsup: Exploiting bounding boxes to supervise convolutional networks
  for semantic segmentation.
\newblock In {\em Proc. IEEE International Conference on Computer Vision
  (ICCV)}, 2015.

\bibitem{imagenet}
Jia Deng, Wei Dong, Richard Socher, Li-Jia Li, Kai Li, and Li Fei-Fei.
\newblock Imagenet: A large-scale hierarchical image database.
\newblock In {\em Proc. IEEE Conference on Computer Vision and Pattern
  Recognition (CVPR)}, 2009.

\bibitem{doersch2015unsupervised}
Carl Doersch, Abhinav Gupta, and Alexei~A Efros.
\newblock Unsupervised visual representation learning by context prediction.
\newblock In {\em Proc. IEEE International Conference on Computer Vision
  (ICCV)}, 2015.

\bibitem{CIAN}
Junsong Fan, Zhaoxiang Zhang, and Tieniu Tan.
\newblock Cian: Cross-image affinity net for weakly supervised semantic
  segmentation.
\newblock {\em arXiv preprint arXiv:1811.10842}, 2018.

\bibitem{DANet}
Jun Fu, Jing Liu, Haijie Tian, Yong Li, Yongjun Bao, Zhiwei Fang, and Hanqing
  Lu.
\newblock Dual attention network for scene segmentation.
\newblock In {\em Proc. IEEE Conference on Computer Vision and Pattern
  Recognition (CVPR)}, 2019.

\bibitem{gidaris2018unsupervised}
Spyros Gidaris, Praveer Singh, and Nikos Komodakis.
\newblock Unsupervised representation learning by predicting image rotations.
\newblock {\em arXiv preprint arXiv:1803.07728}, 2018.

\bibitem{GAN}
Ian Goodfellow, Jean Pouget-Abadie, Mehdi Mirza, Bing Xu, David Warde-Farley,
  Sherjil Ozair, Aaron Courville, and Yoshua Bengio.
\newblock Generative adversarial nets.
\newblock In {\em Proc. Neural Information Processing Systems (NIPS)}, 2014.

\bibitem{SBD}
Bharath Hariharan, Pablo Arbelaez, Lubomir Bourdev, Subhransu Maji, and
  Jitendra Malik.
\newblock Semantic contours from inverse detectors.
\newblock In {\em Proc. IEEE International Conference on Computer Vision
  (ICCV)}, 2011.

\bibitem{SeeNet}
Qibin Hou, PengTao Jiang, Yunchao Wei, and Ming-Ming Cheng.
\newblock Self-erasing network for integral object attention.
\newblock In {\em Proc. Neural Information Processing Systems (NIPS)}, 2018.

\bibitem{DSRG}
Zilong Huang, Xinggang Wang, Jiasi Wang, Wenyu Liu, and Jingdong Wang.
\newblock Weakly-supervised semantic segmentation network with deep seeded
  region growing.
\newblock In {\em Proc. IEEE Conference on Computer Vision and Pattern
  Recognition (CVPR)}, 2018.

\bibitem{SCOPS}
Wei-Chih Hung, Varun Jampani, Sifei Liu, Pavlo Molchanov, Ming-Hsuan Yang, and
  Jan Kautz.
\newblock Scops: Self-supervised co-part segmentation.
\newblock In {\em Proc. IEEE Conference on Computer Vision and Pattern
  Recognition (CVPR)}, 2019.

\bibitem{SDI}
Anna Khoreva, Rodrigo Benenson, Jan Hosang, Matthias Hein, and Bernt Schiele.
\newblock Simple does it: Weakly supervised instance and semantic segmentation.
\newblock In {\em Proc. IEEE Conference on Computer Vision and Pattern
  Recognition (CVPR)}, 2017.

\bibitem{SEC}
Alexander Kolesnikov and Christoph~H Lampert.
\newblock Seed, expand and constrain: Three principles for weakly-supervised
  image segmentation.
\newblock In {\em Proc. European Conference on Computer Vision (ECCV)}, 2016.

\bibitem{larsson2016learning}
Gustav Larsson, Michael Maire, and Gregory Shakhnarovich.
\newblock Learning representations for automatic colorization.
\newblock In {\em Proc. European Conference on Computer Vision (ECCV)}, 2016.

\bibitem{FickleNet}
Jungbeom Lee, Eunji Kim, Sungmin Lee, Jangho Lee, and Sungroh Yoon.
\newblock Ficklenet: Weakly and semi-supervised semantic image segmentation
  using stochastic inference.
\newblock In {\em Proc. IEEE Conference on Computer Vision and Pattern
  Recognition (CVPR)}, 2019.

\bibitem{scribblesup}
Di Lin, Jifeng Dai, Jiaya Jia, Kaiming He, and Jian Sun.
\newblock Scribblesup: Scribble-supervised convolutional networks for semantic
  segmentation.
\newblock In {\em Proc. IEEE Conference on Computer Vision and Pattern
  Recognition (CVPR)}, 2016.

\bibitem{FCN}
Jonathan Long, Evan Shelhamer, and Trevor Darrell.
\newblock Fully convolutional networks for semantic segmentation.
\newblock In {\em Proc. IEEE Conference on Computer Vision and Pattern
  Recognition (CVPR)}, 2015.

\bibitem{EM-Adapt}
George Papandreou, Liang-Chieh Chen, Kevin~P Murphy, and Alan~L Yuille.
\newblock Weakly-and semi-supervised learning of a deep convolutional network
  for semantic image segmentation.
\newblock In {\em Proc. IEEE International Conference on Computer Vision
  (ICCV)}, 2015.

\bibitem{CCNN}
Deepak Pathak, Philipp Krahenbuhl, and Trevor Darrell.
\newblock Constrained convolutional neural networks for weakly supervised
  segmentation.
\newblock In {\em Proc. IEEE International Conference on Computer Vision
  (ICCV)}, 2015.

\bibitem{pathak2016context}
Deepak Pathak, Philipp Krahenbuhl, Jeff Donahue, Trevor Darrell, and Alexei~A
  Efros.
\newblock Context encoders: Feature learning by inpainting.
\newblock In {\em Proc. IEEE Conference on Computer Vision and Pattern
  Recognition (CVPR)}, 2016.

\bibitem{MIL}
Pedro~O Pinheiro and Ronan Collobert.
\newblock From image-level to pixel-level labeling with convolutional networks.
\newblock In {\em Proc. IEEE Conference on Computer Vision and Pattern
  Recognition (CVPR)}, 2015.

\bibitem{GradCAM}
Ramprasaath~R Selvaraju, Michael Cogswell, Abhishek Das, Ramakrishna Vedantam,
  Devi Parikh, and Dhruv Batra.
\newblock Grad-cam: Visual explanations from deep networks via gradient-based
  localization.
\newblock In {\em Proc. IEEE International Conference on Computer Vision
  (ICCV)}, 2017.

\bibitem{selfattention}
Ashish Vaswani, Noam Shazeer, Niki Parmar, Jakob Uszkoreit, Llion Jones,
  Aidan~N Gomez, {\L}ukasz Kaiser, and Illia Polosukhin.
\newblock Attention is all you need.
\newblock In {\em Proc. Neural Information Processing Systems (NIPS)}, 2017.

\bibitem{RAWK}
Paul Vernaza and Manmohan Chandraker.
\newblock Learning random-walk label propagation for weakly-supervised semantic
  segmentation.
\newblock In {\em Proc. IEEE Conference on Computer Vision and Pattern
  Recognition (CVPR)}, 2017.

\bibitem{nonlocal}
Xiaolong Wang, Ross Girshick, Abhinav Gupta, and Kaiming He.
\newblock Non-local neural networks.
\newblock In {\em Proc. IEEE Conference on Computer Vision and Pattern
  Recognition (CVPR)}, 2018.

\bibitem{AdvErasing}
Yunchao Wei, Jiashi Feng, Xiaodan Liang, Ming-Ming Cheng, Yao Zhao, and
  Shuicheng Yan.
\newblock Object region mining with adversarial erasing: A simple
  classification to semantic segmentation approach.
\newblock In {\em Proc. IEEE Conference on Computer Vision and Pattern
  Recognition (CVPR)}, 2017.

\bibitem{STC}
Yunchao Wei, Xiaodan Liang, Yunpeng Chen, Xiaohui Shen, Ming-Ming Cheng, Jiashi
  Feng, Yao Zhao, and Shuicheng Yan.
\newblock Stc: A simple to complex framework for weakly-supervised semantic
  segmentation.
\newblock {\em IEEE Transactionson Pattern Analysis and Machine Intelligence
  (TPAMI)}, 39(11):2314--2320, 2017.

\bibitem{MDC}
Yunchao Wei, Huaxin Xiao, Honghui Shi, Zequn Jie, Jiashi Feng, and Thomas~S
  Huang.
\newblock Revisiting dilated convolution: A simple approach for weakly-and
  semi-supervised semantic segmentation.
\newblock In {\em Proc. IEEE Conference on Computer Vision and Pattern
  Recognition (CVPR)}, 2018.

\bibitem{resnet38}
Zifeng Wu, Chunhua Shen, and Anton Van Den~Hengel.
\newblock Wider or deeper: Revisiting the resnet model for visual recognition.
\newblock {\em Pattern Recognition}, 90:119--133, 2019.

\bibitem{MCOF}
Wang Xiang, You Shaodi, Li Xi, and Ma Huimin.
\newblock Weakly-supervised semantic segmentation by iteratively mining common
  object features.
\newblock In {\em Proc. IEEE Conference on Computer Vision and Pattern
  Recognition (CVPR)}, 2018.

\bibitem{OCNet}
Yuhui Yuan and Jingdong Wang.
\newblock Ocnet: Object context network for scene parsing.
\newblock {\em arXiv preprint arXiv:1809.00916}, 2018.

\bibitem{PSPNet}
Hengshuang Zhao, Jianping Shi, Xiaojuan Qi, Xiaogang Wang, and Jiaya Jia.
\newblock Pyramid scene parsing network.
\newblock In {\em Proc. IEEE Conference on Computer Vision and Pattern
  Recognition (CVPR)}, 2017.

\bibitem{CAM}
Bolei Zhou, Aditya Khosla, Agata Lapedriza, Aude Oliva, and Antonio Torralba.
\newblock Learning deep features for discriminative localization.
\newblock In {\em Proc. IEEE Conference on Computer Vision and Pattern
  Recognition (CVPR)}, 2016.

\end{thebibliography}
	}
	
\end{document}